\documentclass{article}

\usepackage{arxiv}
\usepackage[utf8]{inputenc} % Allow UTF-8 input
\usepackage[T1]{fontenc}    % Use 8-bit T1 fonts
\usepackage{hyperref}       % Hyperlinks
\usepackage{url}            % Simple URL typesetting
\usepackage{booktabs}       % Professional-quality tables
\usepackage{amsfonts}       % Blackboard math symbols
\usepackage{nicefrac}       % Compact symbols for 1/2, etc.
\usepackage{microtype}      % Microtypography
\usepackage{graphicx}       % Graphics handling
\usepackage{natbib}         % Bibliography management
\usepackage{doi}            % Handles DOI
\usepackage{tabularx}       % Extended table functionalities

\newcolumntype{L}{>{\raggedright\arraybackslash}X} % left justified
\newcolumntype{C}{>{\centering\arraybackslash}X}   % centered
\newcolumntype{R}{>{\raggedleft\arraybackslash}X}  % right justified

\title{Evaluating Text Summaries Generated by Large Language Models Using OpenAI's GPT}

\author{
  Hassan Shakil \\
  University of Colorado - Colorado Springs \\
  Colorado Springs, CO \\
  \texttt{hshakil@uccs.edu} \\
  \And
  Atqiya Munawara Mahi \\
  University of Massachusetts Lowell \\
  \texttt{} \\
  \And
  Phuoc Nguyen \\
  University of Kansas \\
  \texttt{phuoc.h.nguyen6.mil@army.mil} \\
  \And
  Zeydy Ortiz \\
  DataCrunch Lab, LLC \\
  Cary, NC \\
  \texttt{zortiz@datacrunchlab.com} \\
  \And
  Mamoun T. Mardini \\
  Department of Health Outcomes and Biomedical Informatics, University of Florida \\
  \texttt{malmardini@ufl.edu} \\
}

\date{}  % This line removes the date

\renewcommand{\shorttitle}{\textit{Evaluating Text Summaries with OpenAI's GPT Models}}

\hypersetup{
pdftitle={Evaluating Text Summaries Generated by Large Language Models Using OpenAI's GPT},
pdfsubject={cs.CL, cs.LG},
pdfauthor={Hassan Shakil, Atqiya Munawara Mahi, Phuoc Nguyen, Zeydy Ortiz, Mamoun T. Mardini},
pdfkeywords={Text Summarization, Evaluation, GPT, ChatGPT, Large Language Models},
}

\begin{document}
\maketitle

\begin{abstract}
This research examines the effectiveness of OpenAI's GPT models as independent evaluators of text summaries generated by six transformer-based models from Hugging Face: DistilBART, BERT, ProphetNet, T5, BART, and PEGASUS. We evaluated these summaries based on essential properties of high-quality summary—conciseness, relevance, coherence, and readability—using traditional metrics such as ROUGE and Latent Semantic Analysis (LSA). Uniquely, we also employed GPT not as a summarizer but as an evaluator, allowing it to independently assess summary quality without predefined metrics. Our analysis revealed significant correlations between GPT evaluations and traditional metrics, particularly in assessing relevance and coherence. The results demonstrate GPT’s potential as a robust tool for evaluating text summaries, offering insights that complement established metrics and providing a basis for comparative analysis of transformer-based models in natural language processing tasks.
\end{abstract}

% keywords can be removed

\section{Introduction}

In the contemporary era characterized by a deluge of data, the intelligence community faces the challenge of information overload, needing to process vast amounts of information swiftly and effectively. The ability to generate succinct, clear, and actionable summaries from diverse data sources is crucial, as it often determines the success of strategic objectives in this information-rich environment. As the demand for systems capable of automating large-scale text summarization without compromising on quality or relevance intensifies, the role of such technologies becomes increasingly critical \cite{liu2019text}.

Text summarization, a pivotal task within Natural Language Processing (NLP), has found widespread application across various domains, including news aggregation and the distillation of extensive documents into manageable summaries. The exponential growth in data underscores the utility of text summarization in enhancing content accessibility and comprehension, thus facilitating more efficient navigation through information landscapes \cite{chouikhi2022deep}.

Recent advancements in NLP have been dominated by transformer models, renowned for their ability to discern intricate textual relationships. Introduced by Vaswani et al., these models utilize self-attention mechanisms to capture contextual relationships within text, representing a significant evolution from traditional sequence-to-sequence approaches \cite{vaswani2017attention}. Despite the uniform architecture, transformer models vary considerably in their specific capabilities and performance characteristics.

This study focuses on the comparative analysis of different transformer-based models for text summarization, utilizing a benchmark dataset. Specifically, we examine the efficacy of Open AI's GPT models in evaluating summaries generated by six prominent transformer models from Hugging Face\footnote{https://huggingface.co/}, including DistilBART \cite{lewis2019bart}\footnote{https://huggingface.co/sshleifer/distilbart-cnn-12-6}, BERT \cite{devlin2018bert}, ProphetNet \cite{qi2020prophetnet}, T5 \cite{raffel2020exploring}, BART \cite{lewis2019bart}, and PEGASUS \cite{zhang2020pegasus}. The evaluation metrics will focus on key attributes of effective intelligence reporting: conciseness, relevance, coherence, and readability. Established evaluation methods such as ROUGE \cite{barbella2022rouge}, Latent Semantic Analysis (LSA) \cite{dumais2004latent}, Flesch-Kincaid \cite{flesch2007flesch}, and Compression Ratio \cite{wang2019compression} will be employed to quantitatively assess the generated summaries.

Innovatively, this study also explores the application of GPT not as a summarizer, but as an evaluator, assessing the summaries independently without predefined metrics. This novel approach aims to juxtapose the quality of summaries generated by different models, leveraging the capabilities of GPT to offer insights into the efficacy of current summarization technologies. Through this research, we aim to bridge the gap between traditional summary evaluation techniques and the cutting-edge developments in AI, exploring how these advancements can further enhance the strategic capabilities of the intelligence community in a data-driven world.

\section{Background and Literature Review}
Text summarization, the process of distilling the most essential information from a text into a condensed form, remains a cornerstone of research in Natural Language Processing (NLP). This technique is widely applicable across various sectors including news reporting, automated report generation, and conversational analysis, demonstrating its versatility and utility in managing extensive information \cite{liu2019text, widyassari2022review}.

Historically, text summarization began with rule-based systems that relied on heuristics, such as selecting sentences containing frequently occurring words, a method pioneered by Luhn \cite{luhn1958automatic}. Although these initial approaches were straightforward, they often failed to grasp the complexities and nuanced meanings of natural language, leading to the exploration of more sophisticated machine learning strategies \cite{yadav2022extractive}.

Among the earliest machine learning techniques applied to text summarization were Recurrent Neural Networks (RNNs), including Long Short-Term Memory (LSTM) units, known for their efficacy in capturing temporal dependencies within text sequences \cite{hochreiter1997long}. Despite their capabilities, RNNs struggled with long sequences, frequently encountering vanishing and exploding gradient issues that hindered performance. In response, Nallapati et al. introduced SummaRuNNer, an innovative extractive summarization model utilizing a hierarchical RNN framework, designed to mitigate these challenges \cite{nallapati2017summarunner}.

The development of transformer models, as conceptualized by Vaswani et al., marked a significant breakthrough in NLP \cite{vaswani2017attention}. Unlike RNNs, transformers leverage a self-attention mechanism, allowing for a broader and more effective capture of contextual relationships across text. The introduction of BERT by Devlin et al. revolutionized text representation through self-supervised training on extensive text corpora, although it was primarily optimized for embedding generation rather than direct text production \cite{devlin2018bert}.

Subsequent innovations led to the creation of models capable of both understanding and generating text, tailored specifically for tasks like summarization. Notably, models such as BART by Lewis et al. and T5 by Raffel et al. have demonstrated exceptional performance in summarization benchmarks due to their robust architecture and training methodologies \cite{lewis2019bart, raffel2020exploring}. Additionally, models like PEGASUS by Zhang et al. and ProphetNet by Yan et al. introduced novel pretraining objectives that further enhanced their summarization capabilities \cite{zhang2020pegasus, qi2020prophetnet}.

Moreover, the advent of knowledge distillation techniques, exemplified by DistilBART, has facilitated the deployment of large transformer models in environments with limited resources, maintaining high performance while reducing computational demands \cite{sanh2019distilbert}.

This study uses the CNN/Daily Mail dataset to assess how well OpenAI's GPT models evaluate text summaries from leading transformer models like DistilBART, BERT, ProphetNet, T5, BART, and PEGASUS. By applying traditional metrics such as ROUGE and Latent Semantic Analysis (LSA) alongside innovative AI-driven evaluations, the research explores the effectiveness of GPT in enhancing the quality of automated text summarization. This approach provides insights into the practical application of AI tools in processing extensive information environments.

\section{Methodology}

\begin{figure}[h]
    \includegraphics[width=14 cm]{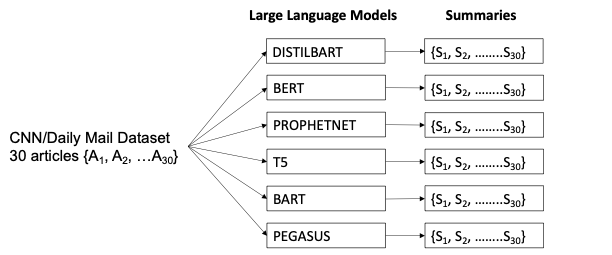}
    \caption{Generation of summaries.\label{Fig_methods}}
\end{figure}

This study rigorously evaluated six transformer-based models, depicted in Figure \ref{Fig_methods}, to assess their efficiency in text summarization tasks. The evaluation was structured around two principal methods: the GPT-Based evaluation and the traditional metrics-based evaluation. The inclusion of the metrics-based method serves as a benchmark to provide a comparative analysis against the GPT-based assessment.

\unskip

Each model selected for this study showcases unique architectural and pre-training innovations that have been demonstrated to enhance performance across various NLP tasks. These models were examined using the widely recognized CNN/Daily Mail dataset, a standard benchmark in the text summarization field, which consists of a substantial compilation of online news articles \cite{hermann2015teaching}. Following are the transformer-based models that we used to generate summaries:

\begin{itemize}

\item \textbf{DistilBART (sshleifer/distilbart-cnn-12-6) \footnote{https://huggingface.co/sshleifer/distilbart-cnn-12-6}} 
DistilBART, a streamlined version of the BART model, employs knowledge distillation, a technique introduced by Hinton et al. This process involves training the distilled model (DistilBART) to emulate the output of the original model (BART), thereby maintaining high performance while enhancing resource efficiency \cite{hinton2015distilling, lewis2019bart}

\item \textbf{BERT-small2BERT-small (mrm8488/bert-small2bert-small-finetuned-cnn\_daily\_mail-summarization) \footnote{https://huggingface.co/mrm8488/bert-small2bert-small-finetuned-cnn\_daily\_mail-summarization/blob/main/README.md?code=true}} This model utilizes a smaller version of BERT, a transformer-based model that introduced the idea of bi-directional training in transformers. BERT has been a milestone in transformer-based models due to its impressive performance in various NLP tasks Devlin et al. \cite{devlin2018bert}

\item \textbf{ProphetNet (microsoft/prophetnet-large-uncased-cnndm) \footnote{https://huggingface.co/microsoft/prophetnet-large-uncased-cnndm}} ProphetNet introduces a novel self-supervised learning objective known as future n-gram prediction, coupled with an n-stream self-attention mechanism. These features enable ProphetNet to predict multiple tokens ahead during pre-training, enhancing its capability for sequence generation tasks \cite{qi2020prophetnet}.

\item \textbf{T5-small (t5-small) \footnote{https://huggingface.co/t5-small}} T5, or Text-to-Text Transfer Transformer, redefines NLP tasks into a text-to-text framework. The 'small' variant of T5 offers a reduction in model size while retaining considerable performance, making it suitable for various applications \cite{raffel2020exploring}.

\item \textbf{BART-large (facebook/bart-large-cnn) \footnote{https://huggingface.co/facebook/bart-large-cnn}} BART integrates the use of a denoising autoencoder for pre-training, distinguishing itself by learning to reconstruct the original text from corrupted versions. This unique approach has proven effective across a multitude of downstream NLP tasks \cite{lewis2019bart}.

\item \textbf{PEGASUS (google/pegasus-cnn\_dailymail)} \footnote{https://huggingface.co/google/pegasus-cnn\_dailymail} PEGASUS employs a unique pre-training objective where specific sentences are omitted, and the model is tasked with regenerating them. This method has shown remarkable effectiveness in abstractive text summarization tasks \cite{zhang2020pegasus}.

\end{itemize}

For the implementation of this study, we utilized the Hugging Face's Transformers library, which offers a suite of pre-trained models and APIs for text processing tasks \cite{wolf2020transformers}. The summarization capabilities of each transformer model were leveraged to process and summarize texts from the CNN/Daily Mail dataset.

We evaluated the performance of these models by analyzing the quality of summaries for selected articles, each limited to 512 words or less. % The results were then organized in a Pandas DataFrame to enable a high-level comparative analysis across the different models \cite{reback2021pandas}.
This setup provided a streamlined method for assessing the effectiveness of each model's summarization technique.

\subsection{GPT-Based Evaluation}
\begin{figure}[h]
\includegraphics[width=14 cm]{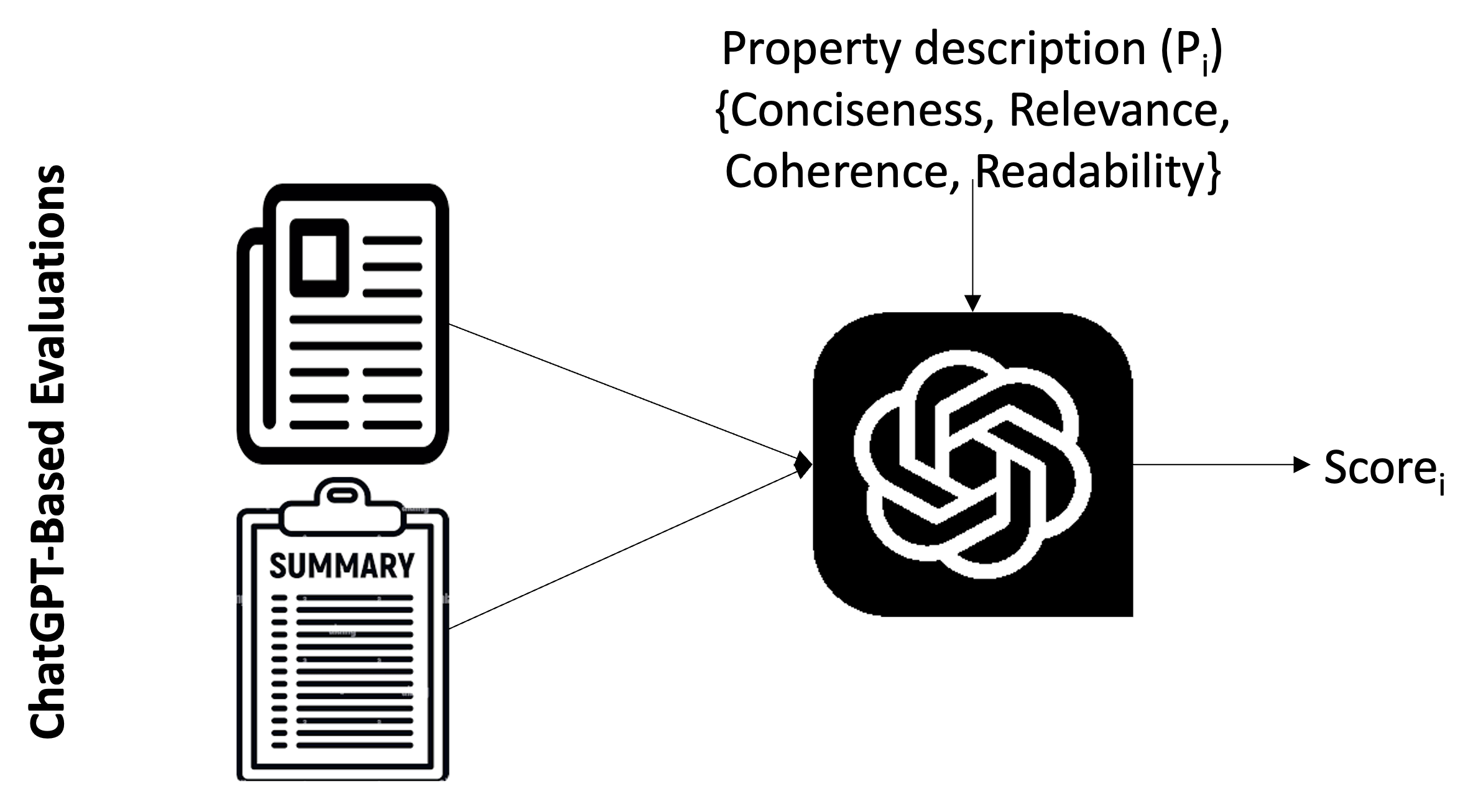}
\caption{Illustration of the GPT-Based evaluation.\label{chatgpt}}
\end{figure}

The evaluation of generated summaries was based on four essential properties of high-quality summaries: conciseness, relevance, coherence, and readability. These properties were aligned with specific comparative benchmarks evaluation metrics, detailed in Table \ref{tab1}. For the evaluation, we utilized the OpenAI API to interact with the GPT-3.5 model, a process illustrated in Figure \ref{chatgpt}. 

\subsubsection{Prompts}
The evaluation involved distinct types of prompts to assess the summaries, each designed to measure different aspects of summary quality. Below is an example of one such prompt used in the study:

\begin{itemize}
\item \textbf{Zero-shot Evaluation:} This prompt type is designed to quickly ascertain the consistency of the summary with the source article. Consistency here refers to the presence and accuracy of information from the source article within the summary.\\

\textbf{Prompt}: Please determine whether the provided summary is consistent with the corresponding article. Note that ``consistency" refers to how much information included in the summary is present in the source article.\\
\textbf{Article}: [Article]\\
\textbf{Summary}: [Summary]\\
\textbf{Answer}: (yes or no)\\

\item \textbf{Chain of thoughts Evaluation:} This approach requires a step-by-step rationale, providing a deeper insight into the decision-making process of the model, enhancing the interpretability of the evaluation.

\textbf{Prompt}: Please determine whether the provided summary is consistent with the corresponding article. Note that ``consistency" refers to how much information included in the summary is present in the source article.\\
\textbf{Article}: [Article]\\
\textbf{Summary}: [Summary]\\
\textbf{Answer}: Explain your reasoning step by step then answer the question (yes or no)\\

\item \textbf{Score Evaluation:} This quantitative approach asks for a direct scoring from 0 to 1, allowing for a precise measurement of consistency, facilitating comparative analysis across different summaries.

\textbf{Prompt}: Score the following summary given the corresponding article with respect to consistency from 0 to 1 where 1 means most consistent.  Note that ``consistency" refers to how much information included in the summary is present in the source article.\\
\textbf{Article}: [Article]\\
\textbf{Summary}: [Summary]\\
\textbf{Score}:\\

\end{itemize}

\begin{table}[h] % Use 'h' for "here" if possible, or use 'ht' for "here or top" as fallback positions
\centering % Centers the table
\caption{Evaluation of the summaries using traditional metrics.}
\label{tab1}
\begin{tabularx}{\textwidth}{XXX} % Three equal columns
\toprule
\textbf{Property} & \textbf{Description} & \textbf{Evaluation metric} \\
\midrule
Conciseness & A high-quality summary should effectively convey the most important information from the original source while keeping the length brief. & Compression Ratio - Calculate the ratio of the length of the summary to the length of the original text. \\
\midrule
Relevance & The information presented in the summary should be relevant to the main topic. & ROUGE (Recall-Oriented Understudy for Gisting Evaluation) - compares n-gram overlap (unigrams, bigrams, etc.) between the summary and the reference summaries or the source text. It assesses how well the summary captures important content. \\
\midrule
Coherence & A good summary should have a clear structure and flow of ideas, making it easy to understand and follow. & Latent Semantic Analysis (LSA) to assess the logical connections between sentences or concepts. \\
\midrule
Readability & The sentence used in the summary should be clear and easily understandable. & Flesch-Kincaid to assess the complexity of sentences in the summary. \\
\bottomrule
\end{tabularx}
\end{table}

\subsection{Metrics-Based Evaluation}
\begin{figure}[h]
\includegraphics[width=14 cm]{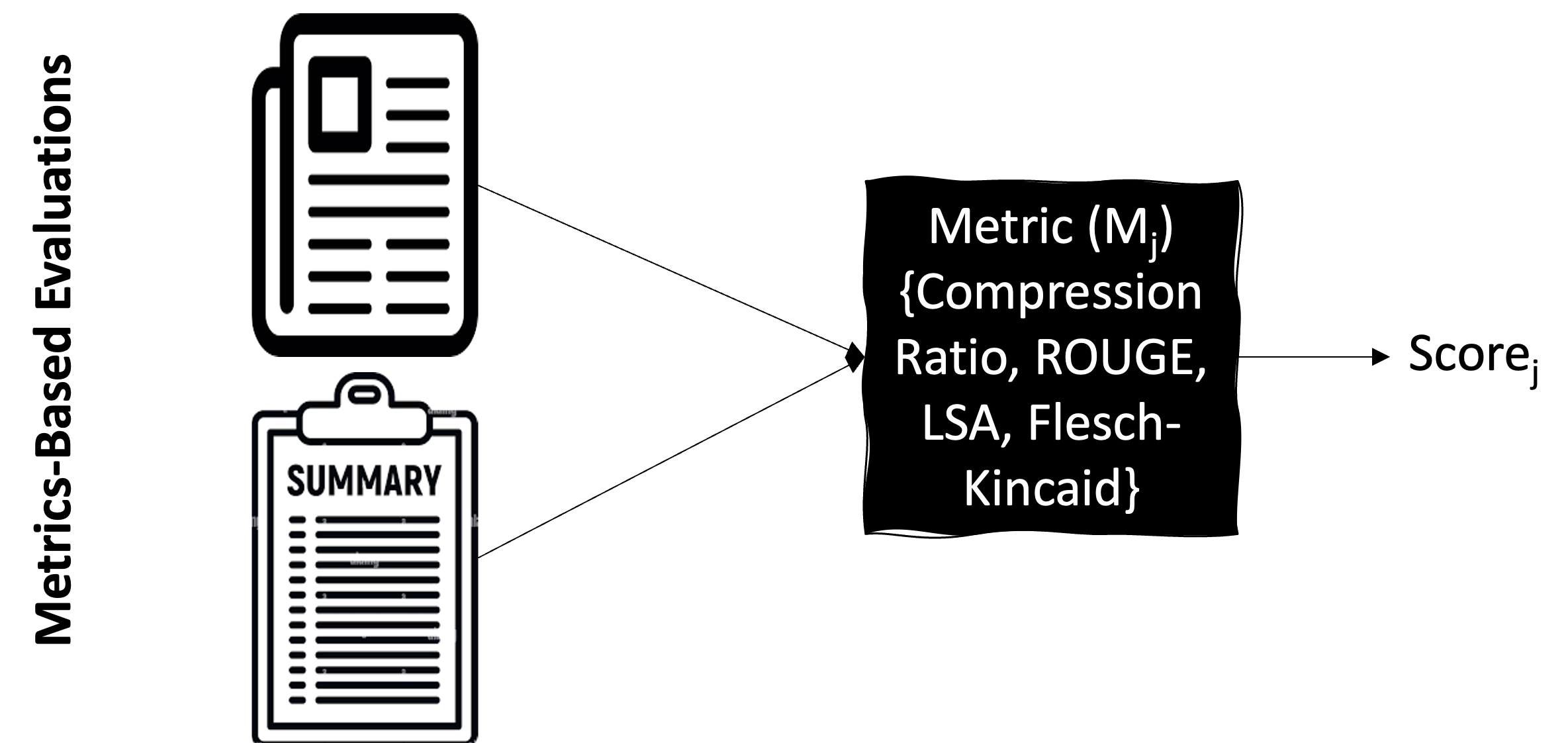}
\caption{Illustration of the Metrics-Based evaluation.\label{metrics}}
\end{figure}

For the metrics-based evaluation of the summaries, we employed several established quantitative metrics to assess their quality. These metrics included the compression ratio, ROUGE (Recall-Oriented Understudy for Gisting Evaluation), Latent Semantic Analysis (LSA), and Flesch-Kincaid readability tests. Each metric provides a unique lens through which the effectiveness of a summary can be measured. The specific calculations for each metric are comprehensively detailed in Table \ref{tab1} and are visually illustrated in Figure \ref{metrics}. This approach allows for a multifaceted assessment of the summaries, highlighting their strengths and areas for improvement in terms of brevity, content fidelity, semantic preservation, and readability.

\subsection{Statistical analysis}
In our study, we undertook a comparative analysis between conventional metrics and the scores generated by GPT. To assess the extent of agreement and association between these two sets of evaluations, we employed Pearson's correlation coefficient \cite{sedgwick2012pearson} a statistical measure widely used to evaluate the linear relationships between two continuous variables.

The primary objective was to ascertain the correlation between the conventional metrics (compression ratio, ROUGE, LSA, and Flesch-Kincaid) and the scores generated by GPT. By executing Pearson's correlation analysis, we aimed to uncover insights into how closely these two methods align in their assessments of summary quality.

The correlation coefficient resulting from Pearson's analysis provides a numerical indicator of the relationship's strength. A higher correlation coefficient indicates a strong positive relationship, suggesting that the evaluations from the conventional metrics and GPT are closely aligned. On the other hand, a lower correlation coefficient would indicate a weaker relationship, highlighting potential discrepancies in how the two methods evaluate summaries. This analysis is crucial for understanding the consistency and validity of the evaluation methods used in our study.

\section{Results}
The results section presents the outcomes of both the metrics-based and GPT-based evaluations of summaries generated by the various Large Language Models (LLMs) studied. These results provide insights into the performance of each model across different evaluative criteria.

\subsection{GPT-Based Evaluation}
Table \ref{tab6} displays the GPT-based scores, which also represent the averages across 30 summaries and are normalized to the 0-1 scale. These scores provide a direct comparison of how well each model performs according to the AI evaluator in terms of the same summary properties.

\begin{table}[h] 
\caption{Scoring the properties with GPT-3.5.\label{tab6}}
\newcolumntype{C}{>{\centering\arraybackslash}X}
\begin{tabularx}{\textwidth}{CCCCCC}
\toprule
\textbf{Model}	& \textbf{Conciseness }	& \textbf{Relevance }& \textbf{Coherence }& \textbf{Readability }\\
\midrule
DISTILBART		& 0.24	& 0.78    & 0.70	& 0.79\\
BERT		& 0.31	& 0.72    & 0.64		& 0.73\\
PROPHETNET		& 0.35	& 0.62    & 0.38	& 0.69\\
T5	& 0.31	& 0.73    & 0.59			& 0.71\\
BART		& 0.26	& 0.81    & 0.72	& 0.82\\
PEGASUS		& 0.24	& 0.79    & 0.68	& 0.78\\
\bottomrule

\end{tabularx}
\end{table}

\subsection{Metrics-Based Evaluation}
The metrics-based scores for each LLM are shown in Table \ref{tab5}. These scores are averages computed from 30 summaries and have been normalized to a range from 0 to 1, where 0 indicates the lowest performance and 1 the highest. Notably, for the compression ratio, we adjusted the scores by subtracting them from 1 to better represent the conciseness as a desirable attribute (lower compression ratio indicating higher conciseness).

\begin{table}[h] 
\caption{Metrics results.\label{tab5}}
\newcolumntype{C}{>{\centering\arraybackslash}X}
\begin{tabularx}{\textwidth}{CCCCCC}
\toprule
\textbf{Model}	& \textbf{Conciseness (Compression Ratio)}	& \textbf{Relevance (ROUGE)}& \textbf{Coherence (LSA)}& \textbf{Readability (Flesch-Kincaid)}\\
\midrule
DISTILBART		& 0.19	& 0.36    & 0.57			& 0.45\\
BERT		& 0.17	& 0.25    & 0.56			& 0.42\\
PROPHETNET		& 0.05	& 0.08    & 0.29			& 0.38\\
T5	& 0.15	& 0.29    & 0.59			& 0.43\\
BART		& 0.16	& 0.33    & 0.57			& 0.40\\
PEGASUS		& 0.13	& 0.28    & 0.49			& 0.38\\
\bottomrule

\end{tabularx}
\end{table}

\subsection{Statistical Analysis of Correlation}
The Pearson's correlation coefficients and P-values, displayed in Table \ref{tab7}, provide statistical insight into the relationships between the metrics-based scores and the GPT-3.5-based scores for each summary property.

\begin{table}[h] 
\caption{Correlation between Metrics-Based and GPT-3.5-Based Evaluations.\label{tab7}}
\newcolumntype{C}{>{\centering\arraybackslash}X}
\begin{tabularx}{\textwidth}{CCC}
\toprule
\textbf{Property}	& \textbf{Correlation coefficient }	& \textbf{P Value }\\
\midrule
Conciseness		& -0.65	& 0.17    \\
Relevance		& 0.92	& 0.01    \\
Coherence		& 0.85	& 0.03    \\
Readability	    & 0.11	& 0.83    \\

\bottomrule

\end{tabularx}
\end{table}

\begin{figure}[h]
\includegraphics[width=14 cm]{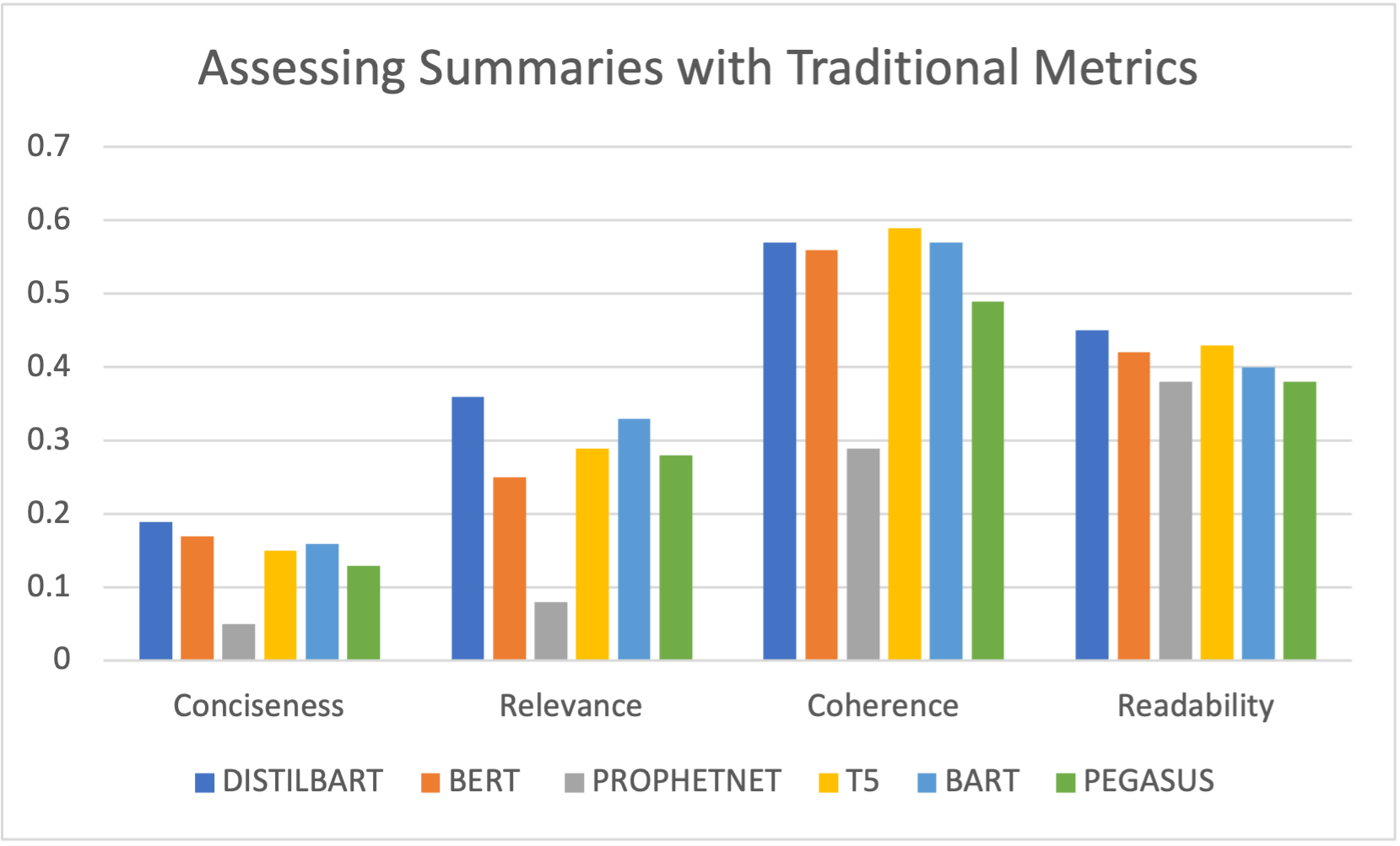}
\caption{Assessing using traditional metrics.\label{fig3}}
\end{figure}

\begin{figure}[h]
\includegraphics[width=14 cm]{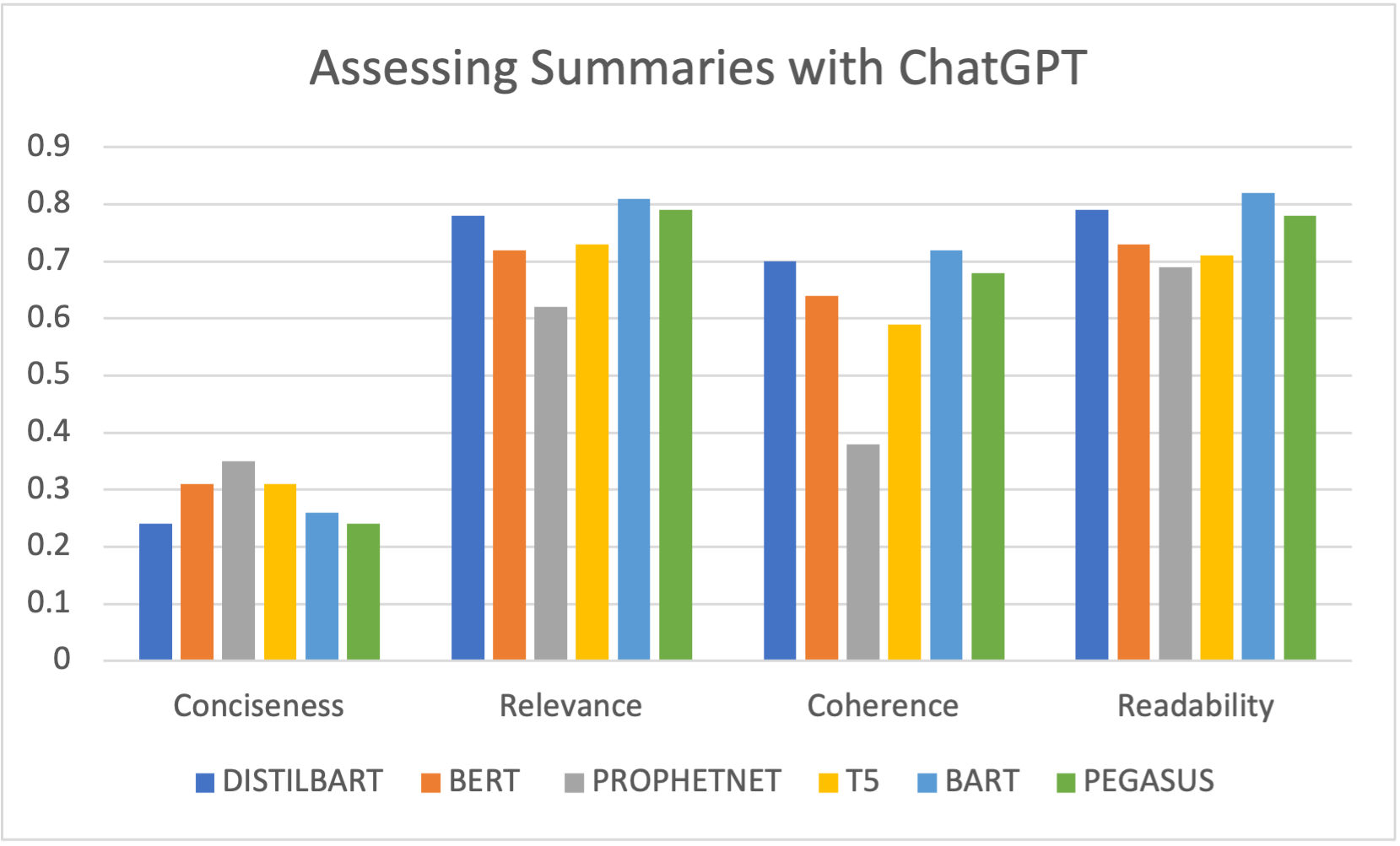}
\caption{Assessing using GPT.\label{fig4}}
\end{figure}

Figures \ref{fig3} and \ref{fig4} visualize the comparative evaluations using traditional metrics and GPT, respectively, highlighting the performance variations across the models.

This comprehensive presentation of results facilitates a nuanced understanding of how each model's performance is perceived differently by conventional evaluation metrics versus AI-driven evaluation, reflecting on the potential alignments and discrepancies between these two assessment methods.

\section{Discussion}

In this study, we employed traditional metrics such as ROUGE and Latent Semantic Analysis (LSA) and introduced GPT as an independent evaluator to assess text summaries. The correlation observed between GPT's assessments and traditional metrics, especially in relevance and coherence, highlights GPT's effectiveness in evaluating these aspects of summaries. Notably, GPT often awarded higher scores than traditional metrics, suggesting it evaluates using a broader set of factors, potentially capturing nuances that traditional metrics overlook.

The performance of the large language models (LLMs) was relatively uniform, except for the PROPHETNET model, which received lower scores from both evaluation types. This can be attributed to PROPHETNET's highly concise outputs, which often failed to provide a comprehensive representation of the original text, underscoring a limitation in capturing detailed content.

These findings demonstrate the utility of integrating AI tools like GPT in the evaluation process, offering a more nuanced perspective compared to traditional metrics alone. This approach not only enriches the evaluation landscape but also helps refine our understanding of model capabilities and limitations in natural language processing tasks.

\section{Conclusion}
In this study, we employed OpenAI's GPT model alongside standard metrics such as ROUGE and LSA to evaluate text summaries. The findings revealed a strong correlation, particularly in the evaluation of relevance and coherence, suggesting that GPT can effectively complement traditional assessment tools. GPT tended to assign higher scores, which likely reflects its ability to consider a broader range of factors in its evaluations. Notably, summaries generated by the PROPHETNET model consistently received lower scores, primarily due to their concise nature and limited content representation. Overall, the results demonstrate GPT's potential as an evaluator and its capacity to enhance traditional metrics, offering valuable insights for future research in natural language processing. This integration of AI-driven tools with established metrics could lead to more comprehensive and nuanced evaluation methods in the field.
\section{Future Work}

Building on the findings of this study, several avenues for future research present themselves. First, expanding the evaluation framework to include more diverse NLP tasks, such as sentiment analysis or entity recognition, could provide a broader understanding of GPT's capabilities as an evaluator across different contexts. Additionally, exploring other transformer-based models not covered in this study could offer insights into how various architectures influence the efficacy of AI-driven evaluation tools.

Another promising direction would be to refine the methodology for integrating AI-driven evaluations with traditional metrics, potentially developing a hybrid model that leverages the strengths of both to produce a more robust evaluation system.

Moreover, there is an opportunity to explore the psychological and perceptual aspects of how humans interpret GPT's evaluations compared to traditional metrics. Understanding these dynamics could lead to improvements in how AI-generated evaluations are presented and utilized, making them more intuitive and actionable for users.

Finally, addressing the limitations observed in models like PROPHETNET, future work could involve developing strategies to enhance the conciseness without sacrificing content comprehensiveness. This could include experimenting with different pre-training and fine-tuning approaches that specifically target the balance between brevity and detail in summary generation.
\bibliographystyle{unsrtnat}
\bibliography{references}  %%% Uncomment this line and comment out the ``thebibliography'' section below to use the external .bib file (using bibtex).

\end{document}